\pdfoutput=1
\documentclass[twoside]{article}

\usepackage[accepted]{aistats2021}
\usepackage{natbib}

\usepackage{times}
\usepackage{helvet}
\usepackage{courier}
\usepackage{amssymb}
\usepackage{amsmath}
\usepackage{amsfonts}
\usepackage[bottom]{footmisc}
\usepackage{wrapfig}
\usepackage{float}
\usepackage[ruled]{algorithm}
\usepackage{algcompatible}
\usepackage{mathtools,nccmath}
\usepackage{amsthm}
\usepackage{amsfonts}
\usepackage{siunitx}
\usepackage{array}
\usepackage{placeins}
\usepackage{hyperref}
\usepackage{url}
\usepackage{xcolor}
\usepackage{multirow}
\usepackage{natbib}
\usepackage{multicol}
\usepackage{booktabs}
\usepackage{graphicx}

\usepackage{amsmath}

\DeclareMathOperator*{\argmax}{arg\,max}

\begin{document}

\newtheorem{theorem}{Theorem}
\newtheorem{definition}{Definition}
\newtheorem{problem}{Problem}
\newtheorem{myexam}{Example}
\newtheorem{assumption}{Assumption}
\newtheorem{proposition}{Proposition}
\newtheorem{remark}{Remark}
\newtheorem{lemma}{Lemma}
\newtheorem{corollary}{Corollary}

\newenvironment{sproof}{%
  \renewcommand{\proofname}{Proof Sketch}\proof}{\endproof}

\newcommand{\MRW}[1]{{\color{blue}[MW: #1]}}
\newcommand{\ap}[1]{{\color{orange}[AP: #1]}}
\newcommand{\LL}[1]{{\color{red}[LL: #1]}}
\newcommand{\MK}[1]{{\color{green}[MK: #1]}}
\newcommand{\ZC}[1]{{\color{pink}[ZC: #1]}}
\newcommand{\ZZ}[1]{{\color{purple}[ZZ: #1]}}
%

%

\twocolumn[
\aistatstitle{Bayesian Inference with Certifiable Adversarial Robustness}

%

\runningauthor{Wicker, Laurenti, Patane, Chen, Zhang, Kwiatkowska}

\aistatsauthor{ Matthew Wicker$^*$ \And  Luca Laurenti$^*$ \And Andrea Patane$^*$}
\aistatsaddress{ Department of Computer Science \\ University of Oxford \And Department of Computer Science \\ University of Oxford \And Department of Computer Science \\ University of Oxford}
\aistatsauthor{  Zhoutong Chen \And Zheng Zhang \And Marta Kwiatkowska}
\aistatsaddress{Department of ECE \\  Univ. of California, Santa Barbara \And Department of ECE  \\  Univ. of California, Santa Barbara \And Department of Computer Science \\ University of Oxford}
]

\begin{abstract}
We consider adversarial training of deep neural networks through the lens of Bayesian learning, and present a principled framework for adversarial training of Bayesian Neural Networks (BNNs) with certifiable guarantees. We rely on techniques from constraint relaxation of non-convex optimisation problems and modify the standard cross-entropy error model to enforce posterior robustness to worst-case perturbations in $\epsilon$-balls around input points.
We illustrate how the resulting framework can be combined with methods commonly employed for approximate inference of BNNs. 
In an empirical investigation, we  demonstrate that the presented approach enables training of certifiably robust models on MNIST, FashionMNIST and CIFAR-10 and can also be beneficial for uncertainty calibration. Our method is the first to directly train certifiable BNNs, thus facilitating their deployment in safety-critical applications.
\end{abstract}

\section{INTRODUCTION}
Although deep neural networks (NNs) have achieved state-of-the-art performance in a range of learning tasks \citep{deepbook}, they have recently been shown to be susceptible to adversarial attacks: small, often imperceptible perturbations to their inputs that can trigger a misclassification \citep{goodfellow6572explaining,biggio2018wild}. 
While retaining the flexibility of standard (deterministic) NNs, Bayesian neural networks (BNNs), i.e., neural networks with distributions placed over their weights and biases, enable principled quantification of their predictions' uncertainty \citep{neal2012bayesian}.
Intuitively, the latter can be used to provide a natural protection against adversarial examples, making BNNs particularly appealing for safety-critical scenarios, in which the safety of the system must be provably guaranteed \citep{mcallister2017concrete}.
In fact, not only have BNNs been shown to possess many favorable robustness properties against adversarial attacks \citep{bekasov2018bayesian,carbone2020robustness}, but their uncertainty has also been investigated as a means of flagging out-of-distribution samples and for robust decision making \citep{kahn17}.  

However, while guarantees for BNNs have been provided for the true Bayesian posterior and under idealised conditions on the training data and network architecture, the necessary assumptions cannot be checked in practice and exact inference for BNNs is generally infeasible. 
Indeed, it has been shown that BNNs trained with modern approximate inference methods and on real-world datasets can be easily fooled by adversarial attacks \citep{grosse2018limitations,athalye2018obfuscated}.
Consequently, BNN methodologies need to be strengthened before they can be deployed in practical safety-critical scenarios. 
However, to the best of our knowledge, 
there is no general Bayesian approach targeted at the training of BNNs with certifiable robustness against adversarial attacks. 

In this paper, we present a principled Bayesian  approach for incorporating adversarial robustness in the posterior inference procedure of BNNs.
In order to do so, we formulate the robustness requirement as the worst-case prediction over an adversarial input ball of radius $\epsilon \geq 0$ induced by a user-defined probability density function $p_\epsilon$, and extend the standard cross-entropy likelihood model by marginalising the network output over $p_\epsilon$.
We refer to the resulting likelihood model as the \textit{robust likelihood}.
We show how, for any $\epsilon > 0$, certified lower bounds to the robust likelihood can be computed by employing Interval Bound Propagation (IBP) techniques \citep{ehlers2017formal,mirman2018differentiable,gowal2018effectiveness}. 
We further demonstrate that, based on a differentiable modification of the likelihood model, the adversarial training procedure we introduce adapts naturally to the main approximate inference techniques employed for training of BNNs, including HMC \citep{neal2011mcmc}, Bayes by Backprop (BBB) \citep{blundell2015weight}, Stochastic Weight Averaging - Gaussian (SWAG) \citep{maddox2019simple}, NoisyAdam (NA) \citep{zhang2018noisy}, and Variational Online Gauss-Newton (VOGN) \citep{osawa2019practical}.

We experimentally investigate the suitability of our method for training certifiably robust BNN models on the MNIST, FashionMNIST and CIFAR-10 datasets. 
We find that training with the robust likelihood enables the first  non-trivial adversarial robustness certification on BNNs.  
In particular, we obtain certified bounds for the robust accuracy of up to $75\%$ on the MNIST test set for $\epsilon = 0.1$, up to $73\%$ on the FashionMNIST test set for $\epsilon = 0.1$, and up to $50\%$ on the CIFAR-10 test set for $\epsilon = 1/255$.
We find the maximum certified safe radius when training with the robust likelihood to be double that obtained when training with the standard likelihood model, which suggests that the robust likelihood pushes the BNN posterior distribution toward more robust regions of the parameter space.
Furthermore, we analyse the effect that the robust likelihood has on the predictive uncertainty. 
We qualitatively observe better calibrated uncertainty when predicting out-of-distribution samples compared with standard training.

In summary, this paper makes the following main contributions: 
\begin{itemize}
    \item We introduce a robust likelihood function based on an adversarial generalisation of the cross-entropy error model for the training of certifiably robust BNNs.
    \item We demonstrate how IBP techniques can be employed to compute a certified lower bound to the robust likelihood 
    for commonly employed approximate Bayesian inference techniques.
    \item We show how our methods allow us to train, for the first time, 
    certified BNNs on MNIST, FashionMNIST and CIFAR-10 with non-trivial robustness. We empirically find that it also leads to improvements in the calibration of uncertainty.\footnote{All of the source code to reproduce the experiments can be found at https://github.com/matthewwicker/CertifiableBayesianInference. All the experiments were run on a single NVIDIA 2080Ti GPU with a 20-core Intel Core Xeon 6230.}
\end{itemize}

\paragraph{Related Works}


Robustness to adversarial examples has been a central topic for both machine learning theorists and practitioners since the seminal work of \cite{IntruigingProperties}. Since their popularization \citep{biggio2018wild}, many methods have been developed to generate adversarial attacks for neural networks~\citep{goodfellow6572explaining, madry2017pgd,carlini2017cwattack} and also specifically for BNNs~\citep{carbone2020robustness,yuan2020gradient}. 
In addition to this, several other methods have been proposed for computing guarantees on the absence of adversarial attacks in the neighbourhood of a given test point  \citep{katz2017reluplex,boopathy2019cnn}, including works that focus on quantifying the robustness to adversarial perturbations of BNNs \citep{cardelli2019statistical, wicker2020probabilistic} and Gaussian processes \citep{cardelli2018robustness,blaas2019robustness}.

Adversarial training, on the other hand, seeks to directly train neural networks that are robust to them \citep{madry2017pgd}. Most of the adversarial training methods have been developed for deterministic (i.e., non-Bayesian) neural networks, where the common approach is that of solving a min-max optimization problem obtained by modifying the loss. 
Based on a similar approach,  \citep{liu2018adv} have developed a method for adversarial training of BNNs trained with Gaussian Variational Inference (VI). However, this method cannot be directly extended to other approximate inference algorithms and relies on gradient-based attacks (i.e., PGD \citep{madry2017pgd}) to approximate worst-case perturbations in the neighbourhood of each data point, which, as we show in Section \ref{sec:experimens}, may fail to generalize to other attacks. Similarly, \cite{ye2018bayesian}, 
assumes a prior distribution around each data point, which is then used to sample adversarial examples. However, by assuming a distribution over the adversarial examples, the resulting approach is not worst-case.
In contrast, the framework presented in this paper is based on a principled Bayesian foundation (i.e., we modify the standard likelihood model to account for adversarial examples). As a consequence, it can be formulated directly for any approximate inference method. Furthermore, by explicitly maximizing a lower bound of the robust likelihood instead of an approximation obtained by employing gradient-based attacks, it allows us to obtain non-trivial certifiable worst-case guarantees for BNNs as illustrated in Section \ref{sec:experimens}.




\section{BAYESIAN INFERENCE WITH NEURAL NETWORKS}\label{sec:background}
%
We consider a generic neural network  (NN) architecture, $f^{w} : \mathbb{R}^n \mapsto \mathbb{R}^C$, parameterised by a vector of weights  $w \in \mathbb{R}^{n_w}$, where $C$ is the number of classes.\footnote{We discuss how to generalise to the regression case in Remark \ref{remark:regression}} 
In a Bayesian setting, one assumes a prior distribution over the weights, $p(w)$, that induces a distribution over the network outputs. 
Given a dataset $\mathcal{D} = \{(x_i, y_i)\}_{i=1}^{n_{\mathcal{D}}}$ our belief over the weight distribution is updated through Bayes' theorem, so as to obtain the posterior weight distribution as $p(w | \mathcal{D}) \propto p(\mathcal{D} | w) p(w) $.
%
%
In the latter equation, $p(\mathcal{D} | w) $  is the data \textit{likelihood} and, under the assumption that data in $\mathcal{D}$ are drawn independently from the same distribution, we have $p(\mathcal{D} | w) = \prod_{i=1}^{n_{\mathcal{D}}} p(y_i|x_i,w) $.
In the standard classification model, the likelihood for a point  $(x_i,y_i)$, i.e., $p(y_i|x_i,w)$, is defined as the multinoulli distribution with the probability for each class given as the softmax of the neural network final logits \citep{bishop1995neural}.
In what follows, we will refer to thus defined $p(y_i|x_i,w)$  as the \emph{standard likelihood}. 
In Section \ref{sec:BayesianAdvTrainingGeneral} we will discuss how the standard likelihood can be modified in order to model adversarial perturbations. 
Given the posterior, $p(\mathcal{D} | w)$, the predictive distribution over a test point $x^*$ is computed as the expected value over the softmax output.
In practice, this is computed via an empirical estimator:
\begin{align}
 \mathbb{E}_{w \sim p(w|\mathcal{D})}&[\sigma(f^{w}(x))] \approx \nonumber\\
 &\hat{\mathbb{E}}^N(x) := \frac{1}{N} \sum_{i=1}^N \sigma \left( f^{w_i}(x) \right), \label{eq:predictor}
\end{align}
\\
where $w_1,\ldots,w_N$ are $N$ random samples from $p(w|\mathcal{D})$ and $\sigma(\cdot)$ is the softmax function.
The output class for a test point $x^*$ is then computed as  $\argmax_{c\in \{1,\ldots,C \}}  \hat{\mathbb{E}}^N(x^*)$.

Given the predictor, a local adversarial example for the BNN at a given test point $x^*$ with associated class label $c^*$ is defined as a point $\bar{x}\in\mathbb{R}^n$ such that $|x^*-\bar{x}| \leq \epsilon$ and $
   \argmax_{c\in \{1,\ldots,C \}} \hat{\mathbb{E}}^N(\bar{x}) \neq c^* ,$    
where $|\cdot|$ is a given  norm, that is, a point in the neighborhood of $x^*$ that gets mis-classified by the BNN predictor.
Given a set of test points $\{x_i^*\}_{i=1,\ldots,m}$, we define the \textit{robust accuracy} (denoted $\mathcal{R}_\epsilon$) for $\epsilon > 0 $ as the ratio between the number of points $x_i^*$ for which no adversarial example exists within $\epsilon$ radius, and the total number of test points, that is: 
\begin{align}
\nonumber     \mathcal{R}_\epsilon = \frac{1}{m} \sum_{i=1}^m  
    \mathbb{I} \bigg[& \forall \bar{x} \; s.t. \; |x^*_i-\bar{x}| \leq \epsilon , \\
    & \argmax_{c\in \{1,\ldots,C \}} \hat{\mathbb{E}}^N(\bar{x}) = c^*_i   \bigg], \label{eq:robust_accuracy}
\end{align}
where $\mathbb{I}[\cdot]$ evaluates to $1$ if the expression inside the brackets is true, and to zero otherwise.

Note that the computation of the predictive distribution relies on the computation of the posterior distribution over the weights, $p(w|\mathcal{D})$.
Exact computation of the latter for BNNs is infeasible and approximate methods are employed for its estimation.
In the following, we briefly describe those employed in this paper.

\paragraph{Hamiltonian Monte Carlo}

Hamiltonian Monte Carlo (HMC) approximates the posterior by defining a Markov chain whose stationary distribution is $p(w | \mathcal{D})$, and relies on 
Hamiltonian dynamics to improve efficiency of the exploration. 
%
This is achieved by alternating between sampling from the potential function $U(w) = -\log(p(w))$, and moving around the weight space by following the dynamics described by a kinetic function, $K(v) = \sum_{i=1}^{n_{w}}{v_i^2}/{(2m_i)}$, given over the auxiliary variable $v$. The hyper-parameters $m_i$ trade-off exploration with exploitation of the weight space \citep{neal2011mcmc}. 

Despite its scalability issues, HMC is considered to be the gold standard of Bayesian inference for neural networks. 
This is due to the fact that, in the limit of the number of simulations of the Markov chain, the HMC approximation converges to the true posterior distribution. 
\paragraph{Variational Inference (VI)} Variational methods assume that the posterior can be approximated by a distribution belonging to a given parametric family, that is: $p(w | \mathcal{D}) \approx q(w |  \theta)$. 
The inference problem then boils down to the that of computing a parameterisation $\theta$ that minimises a given variational objective.
While 
finding the optimal $\theta$ does not guarantee convergence to the true posterior of the BNN, variational methods have significant scalability advantages  \citep{osawa2019practical}. 
A number of variational methods were developed in the literature. 
We briefly review the ones that we employ in this paper below.
\cite{blundell2015weight} introduced Bayes by Backprop (BBB), a stochastic gradient method to update the parameters of the variational distribution using KL divergence.
VI methods with improved scalability have been obtained by relying on the natural gradient (i.e.\ the gradient w.r.t.\ sampled weights) in order to update the variational parameters.
This, in conjunction with momentum, has been shown in many instances to be an efficient method for performing VI at scale, as implemented by VOGN \citep{lin2020handling} and NA \citep{zhang2018noisy}. 
A method presented by \cite{maddox2019simple} (SWAG) generates a variational distribution by moment matching a Gaussian distribution to stochastic gradient descent iterates. 

\section{A BAYESIAN APPROACH FOR ADVERSARIAL TRAINING}
\label{sec:BayesianAdvTrainingGeneral}
In this section we present a method for the adversarial training of BNNs that relies on embedding adversarial robustness in the standard cross-entropy error model.
We first introduce the following notation.
Given a data point $(x,y)$, with $x\in \mathbb{R}^n$ and $y$ an associated class label, an $\epsilon>0$, and a fixed weight $w\in \mathbb{R}^{n_w}$, we define ${f}^{w,\epsilon}_{\min}(x)$ 
to be the vector of logits corresponding to the minimizer of the softmax of class $y$ for any input point in an $\epsilon-$ball around $x$: 

\begin{align}\label{eq:sigma_min_def}
\sigma_{y}({f}^{w,\epsilon}_{\min}(x))= \min_{x':|x-x'|\leq \epsilon} \sigma_y(f^w(x')),
\end{align}

where $\sigma_{y}(\cdot)$ is the softmax relative to class $y.$
In the robust training scenario we would like the likelihood $p(y|x,w)$  
to be influenced not only by the accuracy of the prediction, but also by the robustness of the BNN in the neighbourhood of a point.
We model this by assuming that the likelihood of a prediction, given an input point and a weight vector, also depends on the worst-case perturbation in an $\epsilon-$ball around the input point. 
In particular, we assume that $\epsilon$ is sampled from a non-negative random variable  with  distribution $p_{\epsilon}$, and for each $\epsilon$ in the support of $p_{\epsilon}$ we consider the softmax likelihood computed for 
${f}^{w,\epsilon}_{\min}(x)$. 
Then, by marginalizing over $p_\epsilon$, we obtain the following likelihood function, which we call \emph{robust likelihood}:
\begin{align}
\nonumber
p(y|x,w) =&  \int_{\mathbb{R}_{\geq 0}}  \sigma_{y}({f}^{w,\epsilon}_{\min}(x)) p_{\epsilon}(\epsilon)  d\epsilon\\
=& \mathbb{E}_{\epsilon \sim p_{\epsilon}}[ \sigma_{y}({f}^{w,\epsilon}_{\min}(x))].
\label{Eqn:likelihood}
\end{align}
That is, in the robust likelihood for each $\epsilon$ we consider the worst-case of the standard likelihood for all the points in an $\epsilon$-ball around $x$ (note that $\sigma_{y}({f}^{w,\epsilon}_{\min}(x)$ is a monotonically decreasing function of $\epsilon$) and then compute the average with respect to $p_{\epsilon}.$  We should  stress that Eqn.~\eqref{Eqn:likelihood} defines a marginal probability, and hence is  a well defined probability. 
Furthermore, we also note that for $p_{\epsilon}=\delta_{0}$, the delta function centered in the origin, we recover the standard cross-entropy likelihood model. 

In the particular case of 
deterministic NNs, adversarial training is generally implemented by balancing between adversarial robustness at a fixed $\epsilon > 0$ and accuracy (that is, at $\epsilon=0$) \citep{goodfellow6572explaining, lakshminarayanan2017simple}.
For $0\leq \lambda \leq 1$ and $\eta > 0$, this can be obtained in our setting by considering the following discrete distribution for $\epsilon$:
\begin{align}
    p_{\epsilon}(\epsilon)=\begin{cases}
    \lambda      & \quad \text{if } \epsilon=0\\
    1-\lambda      & \quad \text{if } \epsilon=\eta
  \end{cases}.
  \label{Eqn:discreteDistr}
\end{align}
This leads to the following form for the robust likelihood:
\begin{align*}
  &  p(y|x,w) = \lambda \cdot  \sigma_y(f^w(x)) +  (1-\lambda) \cdot \sigma_y({f}_{\min}^{w,\eta}(x)),
\end{align*}
which is a weighted sum of two softmax functions, one given by the standard likelihood and the other accounting for adversarial robustness. 

By assuming the statistical independence of the training labels given input and weights 
(which is the standard assumption for classification \citep{bishop1995neural}), we obtain the following negative log-likelihood for our model:
%
%
\begin{align}\label{eq:error_model_robust}
 E=-\sum_{i=1}^{n_{\mathcal{D}}} log\big( \mathbb{E}_{\epsilon \sim p_{\epsilon}}[ \sigma_{y^{(i)}}({f}_{\min}^{w,\epsilon}(x^{(i)}))] \big).
\end{align}
Notice that this has a trivial absolute minimum when $\mathbb{E}_{\epsilon \sim p_{\epsilon}}[ \sigma_{y^{(i)}}({f}_{\min}^{w,\epsilon}(x^{(i)}))] =1$ for all $(x^{(i)},y^{(i)})\in \mathcal{D}$.
Hence, the absolute minimum of the negative log-likelihood (which would correspond to the maximum likelihood estimation) is reached for the set of weights $w^*$, if it exists, such that for any $(x^{(i)},y^{(i)})\in \mathcal{D}$ almost surely $f^{w^*}(x^{(i)})$ has no adversarial examples in an $\epsilon-$ball around $x^{(i)}$, for any $\epsilon$ in the support of $p_{\epsilon}$.


Therefore, in order to evaluate our robust likelihood model we need to be able to compute $\sigma_{y}({f}^{w,\epsilon}_{\min}(x))$  as defined by Eqn.~\eqref{eq:sigma_min_def}. 
This is discussed in the next section. 
First, we briefly discuss the case of regression models.

\begin{remark}\label{remark:regression}
The above analysis concerns a classification framework. For a regression problem everything follows similarly except that the likelihood is a Gaussian distribution with variance $\Sigma$ \citep{bishop1995neural}. In particular, assuming for simplicity and  without lost of generality that $C=1$ --  i.e., single output regression -- call 
    $f^{w,\epsilon}_{\max}(x)=\max_{x' : |x-x'|\leq \epsilon} f^{w}(x')$ and $f^{w,\epsilon}_{\min}(x)=\min_{x' : |x-x'|\leq \epsilon} f^{w}(x')$.
That is, $f^{w,\epsilon}_{\max}(x)$ and $f^{w,\epsilon}_{\min}(x)$ are the maximum and minimum of $f^w$ for all the points in an $\epsilon-$ball centered around $x$.
Then, in the regression case, for $x \in \mathbb{R}^{n}, y \in \mathbb{R}$ the robust likelihood is:
\begin{align*}
   & p(y|x,w) = \frac{1}{\sqrt{2\pi\Sigma}}exp\big(-\frac{1}{2\Sigma}\max \{\\
   &\quad (\mathbb{E}_{\epsilon \sim p_{\epsilon}}[f^{w,\epsilon}_{\max}(x)]-y)^2, (\mathbb{E}_{\epsilon \sim p_{\epsilon}}[f^{w,\epsilon}_{\min}(x)]-y)^2\}\big).
\end{align*}  

\end{remark}

\section{ADVERSARIAL TRAINING THROUGH INTERVAL BOUND PROPAGATION}
\label{sec:IBP}
We now describe our implementation of the robust likelihood introduced in Section \ref{sec:BayesianAdvTrainingGeneral}. For a data point $(x,y)$, we need to compute $\sigma_{y}({f}^{w,\epsilon}_{\min}(x)),$ that is, the minimum of the softmax of class $y$ for all the points in an $\epsilon-$ball around $x$.
%
Notice that, the likelihood function, $p(y|x,w)$, is defined over given values of the weight vector $w$ (and input point $x$), so that, even in the Bayesian settings, we need to deal with a single deterministic NN (sampled from the BNN) at a time.
Unfortunately still, the computation of this quantity poses an NP-complete problem for deterministic NNs \citep{katz2017reluplex}.
In this section, we review how Interval Bound Propagation (IBP) can be used to compute a safe lower bound on $\sigma_{y}({f}^{w,\epsilon}_{\min}(x))$ \citep{ehlers2017formal,mirman2018differentiable,gowal2018effectiveness} and then discuss extensions needed to provide certification for BNN posterior predictions. 

\paragraph{Interval Bound Propagation}
IBP works by propagating bounding boxes through the network's layers.
Given a weight vector $w$, 
the computation through the network layers on an input point $x$ is denoted by:
\begin{align*}
    &f^{w}_{(k)} = W^w_{k-1} \phi^{w}_{(k-1)} + b^w_{k-1} \quad k=1,\ldots,K \\
  &\phi^{w}_{(k)} = h(f^{w}_{(k)}) \qquad \qquad \qquad \;  k=1,\ldots,K-1,
\end{align*}
with $\phi^{w}_{0} = x,$
where $K$ is the number of layers, $W^w_{k}$ and $b_k^w$ are the weight matrix and bias vector corresponding to the weight vector $w$ for the $k$th layer of the network, $h$ is the activation function, $f^{w}_{(k)}$ is the $k$th pre-activation and $\phi^{w}_{(k)}$ is the $k$th activation vector.
Notice that $f^{w}_{(K)}$ is the network latent vector, whose softmax provides the final output of the network. 
Following the procedure discussed by \cite{gowal2018effectiveness}, we consider an $\ell_\infty-$ball of radius $\epsilon$ in the  input space that we denote as $[\phi^{w,L}_{0},\phi^{w,U}_{0}]$.
We then iteratively propagate the latter through each layer, for $k=1,\ldots,K$.
This can be done efficiently by the introduction of the auxiliary variables, $\hat{\mu}_{k}, \mu_{k}, \hat{r}_{k}$ and $r_{k}$, as it follows: 
\begin{align*}
      &\hat{\mu}_{k} = \dfrac{\phi^{w,U}_{(k-1)} + \phi^{w,L}_{(k-1)}}{2}, \qquad
  \hat{r}_{k} = \dfrac{\phi^{w,U}_{(k-1)} - \phi^{w,L}_{(k-1)}}{2} \\
  &\mu_{k} = W_{k-1}^w \hat{\mu}_{k} + b_{k-1}^w, \qquad
    r_{k} = | W_{k-1}^w  | \hat{r}_{k}  \\
  & \qquad f^{w,U}_{(k)} = \mu_{k} + r_{k},  \qquad  
  f^{w,L}_{(k)} = \mu_{k} - r_{k} \\
  & \qquad \phi^{w,U}_{(k)} = h(f^{w,U}_{(k)}),  \qquad  
  \phi^{w,L}_{(k)} = h(f^{w,L}_{(k)}).
\end{align*}
The resulting latent values for $k=K$, i.e., $f^{w,U,\epsilon} := f^{w,U}_{(K)}$ and $f^{w,L,\epsilon} := f^{w,L}_{(K)}$, yield a valid bounding box for the network output, that is, they are such that $f^w(x) \in [f^{w,L,\epsilon}, f^{w,U,\epsilon}   ]$ for every $x$ in the considered $\epsilon-$ball \citep{gowal2018effectiveness}.

\paragraph{Bounding the Robust Likelihood}
Consider a generic input point $x$ with associated class $y$, and an $\epsilon > 0$ from the support of $p_\epsilon$.
Let $f^{w,L,\epsilon}$ and $f^{w,U,\epsilon}$ be the extrema of the bounding box computed by IBP as described in the previous paragraph.
For $j=1,\ldots,C$, we define the vector $f^{w,\epsilon}_{LB}(x)$ as  follows:
\begin{align}\label{eq:ibp_lb_logits}
    f^{w,\epsilon}_{LB,j}(x) = \begin{cases} f^{w,U,\epsilon}_j \quad \text{if} \; j \neq y \\
                                       f^{w,L,\epsilon}_j \quad \text{if} \; j = y.
                                     \end{cases}
\end{align}
Since the softmax function for the class $y$ is monotonically increasing along the $y$ coordinate and monotonically decreasing along any other direction, we have that $\sigma_{y}(f^{w,\epsilon}_{LB}(x)) \leq \sigma_{y}({f}^{w,\epsilon}_{\min}(x))$.
That is, $\sigma_{y}(f^{w,\epsilon}_{LB}(x))$ provides a lower bound for $\sigma_{y}({f}^{w,\epsilon}_{\min}(x))$.
By propagating this bound through the computation of the robust likelihood, we obtain the following proposition
\begin{proposition}\label{lemma:ibp_likelihood}
Consider a point $x$ with class $y$, and a distribution over $\epsilon$, $p_\epsilon$. 
Given a weight vector $w$ we have that:
\begin{align*}
    p(y|x,w) \geq \mathbb{E}_{\epsilon \sim p_{\epsilon}}[ \sigma_{y}({f}^{w,\epsilon}_{LB}(x))] := p_{\text{IBP}}(y|x,w).
\end{align*}\label{prop:IBPLikelhiood}
\end{proposition}
\vspace{-1em}
\begin{sproof}
For any $\epsilon$ in the support of $p_\epsilon$ we have, by the construction described above, that $\sigma_{y}({f}^{w,\epsilon}_{\min}(x))  \geq   \sigma_{y}(f^{w,\epsilon}_{LB}(x)) $. 
Hence, since expected values respect inequalities, we have that:
\begin{align*}
    p(y|x,w) &= \mathbb{E}_{\epsilon \sim p_{\epsilon}}[ \sigma_{y}({f}^{w,\epsilon}_{\min}(x))] \geq  \mathbb{E}_{\epsilon \sim p_{\epsilon}}[ \sigma_{y}(f^{w,\epsilon}_{LB}(x)) ] \\
    &=  p_{\text{IBP}}(y|x,w).
\end{align*}
\end{sproof}
Proposition \ref{lemma:ibp_likelihood} guarantees that the robust likelihood function can be lower bounded by using IBP computations.
This provides us with a worst-case analysis for adversarial attacks, in the sense that if the network is safe according to IBP then it is necessarily safe (while the converse is not true).
In fact, while $p(y|x,w)$ could be approximated by means of adversarial attack methods (e.g., FGSM or PGD), the resulting approximation would not have any guarantees with respect to the actual values of the $p(y|x,w)$.
Since $p_{\text{IBP}}(y|x,w)$ is non-negative and a formal lower bound to $p(y|x,w)$, by employing it we obtain a (non-negative) lower bound to Equation \eqref{eq:error_model_robust} as well, 
so that the property of absence of adversarial examples at the optimal value of $p(y|x,w)$ is maintained by $p_{\text{IBP}}(y|x,w)$.
%
However, the same does not hold for an approximation of $p(y|x,w)$ given by adversarial attacks, similarly to how maximising a lower bound yields guarantees on the final result, whereas nothing can be said about the maximisation of an upper bound.
In Section \ref{sec:experimens} we empirically compare IBP with  PGD and find these observations to be confirmed in practice.

\paragraph{Certification of BNNs} 
After training, we can employ IBP to compute the certified robust accuracy of the BNN on a test set.
In order to do so, instead of performing IBP on the single extracted network from the BNN, we have to propagate the input bounding box through the posterior predictor. 
To see that, consider a predictor for the BNN output $\hat{\mathbb{E}}^N$ defined as in Eqn.~\eqref{eq:predictor}. 
Let $w_1,\ldots,w_N$ be the sampled weights related to the latter, consider a given test point $x$ with true label $y$, and let $\epsilon > 0$ be the adversarial perturbation strength that we want to provide certification against.
By performing IBP $N$ times we obtain a family of lower bounds such that $\sigma_{y}(f^{w_i,\epsilon}_{LB}(x)) \leq \sigma_{y}({f}^{w_i,\epsilon}_{\min}(x))$, for $i=1,\ldots,N$. 
By averaging these out we obtain a lower bound on the given empirical predictor 
for class $y$ in $x$, which we denote as $\hat{\mathbb{E}}^{N,\epsilon}_{y,LB} := \frac{1}{N} \sum \sigma_{y}(f^{w_i,\epsilon}_{LB}(x))$. 
Similarly, we can proceed to compute an upper bound on the predictor for classes $c' \neq y$.
For this purpose we define:
\begin{align*}
    f^{w,\epsilon,c'}_{UB,j}(x) = \begin{cases} f^{w,L,\epsilon}_j \quad \text{if} \; j \neq c' \\
                                       f^{w,U,\epsilon}_j \quad \text{if} \; j = c',
                                     \end{cases}
\end{align*}
which we average out to obtain an upper bound on the predictive probability for class $c'$ as $\hat{\mathbb{E}}^{N,\epsilon}_{c',UB} := \frac{1}{N} \sum \sigma_{c'}(f^{w_i,\epsilon,c'}_{UB}(x))$.
By defining the worst-case predictor for the true class $y$ according to IBP for $x$ as the vector $\hat{\mathbb{E}}^N_{IBP,\epsilon}(x) \in \mathbb{R}^C$ such that its $y-$th entry is equal to $ \hat{\mathbb{E}}^{N,\epsilon}_{y,LB} $, while the other entries for $c' \neq y$ are set to $ \hat{\mathbb{E}}^{N,\epsilon}_{c',UB}$, we obtain the following theorem.
\begin{theorem}\label{th:certification}
Consider a point $x^*$ with associated class $y^*$.
Assume that $\argmax_{c \in \{1,\ldots,C\} } \hat{\mathbb{E}}^N_{IBP,\epsilon}(x^*) = y^*$, then:
\begin{align*}
    \argmax_{c \in \{1,\ldots,C\} } \hat{\mathbb{E}}^N(\bar{x}) = y^* \quad \forall \bar{x} \; \; \text{s.t.} \; \; |x^*-\bar{x}| \leq \epsilon.
\end{align*}
As a consequence, given a set of test points $x_i^*$ with associated class labels  $y_i^*$, for $i=1,\ldots,m$, we have the following certified bound for the robust accuracy at $\epsilon > 0$:
\begin{align*}
\mathcal{R}_\epsilon \geq \frac{1}{m}\sum_{i=1}^m \mathbb{I}\left[ \argmax_{c \in \{1,\ldots,C\} } \hat{\mathbb{E}}^N_{IBP,\epsilon}(x_i^*) = y_i^*  \right]:=\mathcal{R}_\epsilon^{IBP}.
\end{align*}
\end{theorem}
\begin{sproof}
By construction we have that,  $\forall \bar{x} \; \; \text{s.t.} \; \; |x^*-\bar{x}| \leq \epsilon$ it holds that:
\begin{align*}
    \hat{\mathbb{E}}^N_{IBP,\epsilon}(x^*)_{y^*}  &\leq \hat{\mathbb{E}}^N_{\epsilon}(\bar{x})_{y^*} , \quad
     \hat{\mathbb{E}}^N_{IBP,\epsilon}(x^*)_{c}  &\geq \hat{\mathbb{E}}^N_{\epsilon}(\bar{x})_{c},
\end{align*}
for $c \neq y^*$, $c \in \{1,\ldots,C\}$.
Hence, if the $\argmax$ of $\hat{\mathbb{E}}^N_{IBP,\epsilon}(x^*)$ is $y^*$, then necessary the $\argmax$ of $\hat{\mathbb{E}}^N_{\epsilon}(\bar{x})$ will be $y^*$ as well (of course, the converse is not necessarily true).
This can be re-stated as:
\begin{align*}
     &\mathbb{I} \bigg[ \forall \bar{x} \; s.t. \; |x^*_i-\bar{x}| \leq \epsilon , \argmax_{c\in \{1,\ldots,C \}} \hat{\mathbb{E}}^N(\bar{x}) = c^*_i   \bigg]  \\
     \geq&  \mathbb{I}\left[ \argmax_{c \in \{1,\ldots,C\} } \hat{\mathbb{E}}^N_{IBP,\epsilon}(x_i^*) = y_i^*  \right].
\end{align*}
The theorem statement then follows by the definitions of $\mathcal{R}_\epsilon $ and $\mathcal{R}_\epsilon^{IBP}$.
\end{sproof}
Theorem \ref{th:certification} 
ensures that a lower bound on the certified robust accuracy of the BNN can be computed by employing IBP on a given predictor.
Observe that, while the guarantees when computing the likelihood in Proposition \ref{lemma:ibp_likelihood} are given for a single weight realisation, so that robust training has the effect of penalising at inference time the weights which are not robust, the certified robust accuracy is computed for an empirical average of predictions over weights sampled from the posterior.
We remark that, while in this paper we only considered training with IBP, linear bound propagation techniques (LBP) \citep{zhang2018efficient} could be similarly employed. 
LBP could potentially yield better bounds at the price of a less efficient training process. 
\begin{remark}
The certified bounds are provided for a given empirical predictor, and as shown in Theorem \ref{th:certification}, if the empirical predictor is known, the bounds are exact.
In the case in which the set of weights $w_1, \ldots, w_N$ changes at every prediction, one can rely on standard concentration inequalities, such as Chernoff's bound, to obtain confidence intervals over the certified behaviour \citep{vapnik2013nature}.
\end{remark}

\begin{figure*}[ht]
\centering
\includegraphics[width=0.92\textwidth]{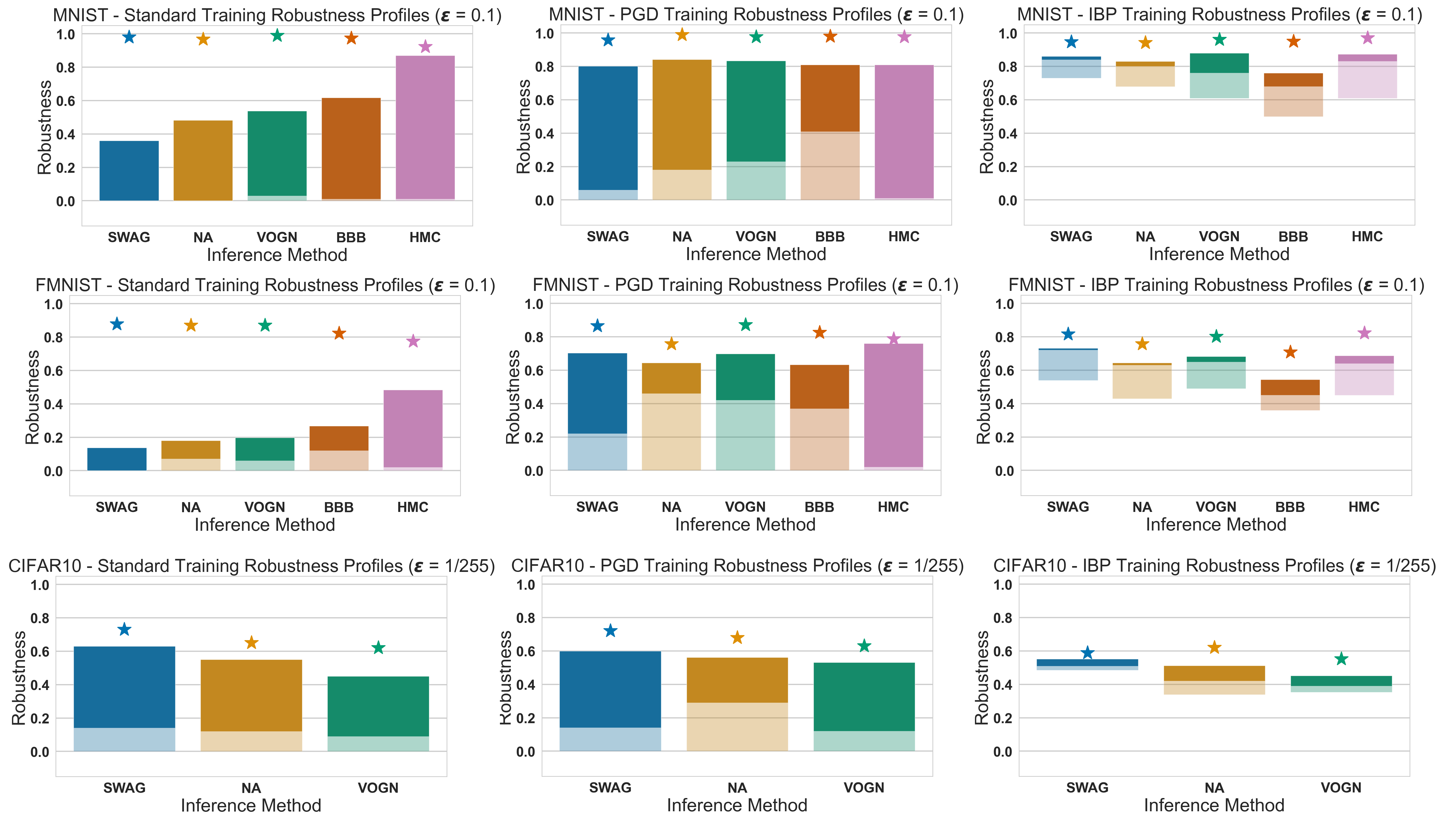}
\caption{
Accuracy (plotted as star points), an empirical estimation of $\mathcal{R}_{\epsilon}$ obtained using PGD (upper bound of each bar), $\mathcal{R}_{\epsilon}^{LBP}$ (lower bound of each bar), and $\mathcal{R}_{\epsilon}^{IBP}$ (shaded lower bound of each bar) obtained for $\epsilon = 0.1$ on the MNIST dataset (top row) and FMNIST (middle row) as well as for $\epsilon = 1/255$ on the CIFAR-10 dataset (bottom row). Each bar refers to a different approximate Bayesian inference technique.
\textbf{Left Column:} results for the standard likelihood. 
\textbf{Centre Column:} results for approximation of robust likelihood using PGD.
\textbf{Right Column:} results for training with
formal IBP lower bound of robust likelihood (Eq~\eqref{eq:error_model_robust}). With our method we obtain up to $75\%$ certified robust accuracy on MNIST and up to $50\%$ on CIFAR-10.
}\label{fig:MNIST}
\end{figure*}

\subsection{APPROXIMATE INFERENCE WITH IBP}
As mentioned in Section \ref{sec:background}, unfortunately, training BNNs is an intractable problem, and remains so when the robust likelihood is used, so that the exact posterior cannot be computed.
Hence, even if a model that perfectly minimises Equation \eqref{eq:error_model_robust} were to exist, we would not be guaranteed to find it (similarly to why no optimality can be claimed when training with the standard likelihood). 
Crucially, however, the likelihood bound computed by IBP is differentiable. 
%
This allows us to adapt commonly used approximate Bayesian inference methods to the robust likelihood settings.
We now give the details for selected variational and Monte Carlo methods. 
We remark that, when computing the certified robust accuracy, we assume that the model has already been trained, so that the inference method does not introduce any error in the certification that we provide.
\paragraph{ROBUST VI}
In Algorithm~\ref{alg:RobustVI} we highlight the changes needed for the implementation of the robust likelihood in the case of natural gradient variational inference.
We maintain the standard form of VI \citep{lin2020handling}, except for lines 6-9, highlighted in red, that emphasize the changes needed in the case in which $p_{\epsilon}$ takes the form of  Eqn.~\eqref{Eqn:discreteDistr}. 
The case of a general $p_{\epsilon}$ can be tackled by iteratively sampling from the  distribution (i.e., by adding a for loop around lines 6-9), the cost of this will, however, be an increased computational time.
Notice, that since the IBP bound is differentiable, so is $l$ defined in line 9 of the algorithm.
We remark that, by changing the parameter update on line 10 with approximations to the Hessian, computing the gradient wrt $\mu, s$, or by introducing momentum parameters, this algorithm can be converted to any of the gradient and natural gradient variational inference methods which have been proposed in recent years, including those of \cite{graves2011practical, blundell2015weight, khan2018fast} and \cite{osawa2019practical}. 

\begin{algorithm}[H]\normalsize
\caption{Robust Natural Grad. Variational Inference}\label{alg:RobustVI}
\textbf{Input:} Prior mean and precision: $\mu_{\text{prior}}, s_{\text{prior}}$, NN architecture: $f$, Dataset: $\mathcal{D}$, Learning rate: $\alpha$, Iterations: $T$, Mini-batch Size: $m$,  $\epsilon$ and $\lambda$ parameters of $p_\epsilon$.\\
\textbf{Output:} Mean and precision of Gaussian approximate posterior.\\
\vspace*{-0.35cm}
\begin{algorithmic}[1]
\STATE $s \gets s_{\text{prior}}$; $\mu \gets \mu_{\text{prior}}$
\FOR{$t = 1,\ldots,T$}
    \STATE $\{X, Y\} \gets \{x_i, y_i\}_{i=0}^{m}$ \Comment{Sample Batch}
    \STATE $w = \mu + ((n_{\mathcal{D}  } s)^{-1/2}  \mathcal{N}(0,I))$
    \STATE ${Y}_{\text{clean}} \gets \sigma(f^w(X))$ 
    \STATE \textcolor{red}{$f^{w,L,\epsilon}(X), f^{w,U,\epsilon}(X) \gets \text{IBP}(f, w, X)$}
    \STATE \textcolor{red}{$f^{w,\epsilon}_{LB}(x) \gets$ Eqn.~\eqref{eq:ibp_lb_logits} for $f^{w,L,\epsilon}(X), f^{w,U,\epsilon}(X) $  } 
    \STATE \textcolor{red}{$Y_{\text{worst}} \gets \sigma(f^{w,\epsilon}_{LB}(x))$}
    \STATE \textcolor{red}{
            $l \gets - Y  log (\lambda Y_{\text{clean}} + (1-\lambda)Y_{\text{worst}})$}
    \Statex \qquad \qquad \qquad  \textcolor{red}{ $+  \mathbb{D}_{\text{KL}}(\mathcal{N}(\mu_{\text{prior}},1/s_{\text{prior}}) \; | \; \mathcal{N}(\mu, 1/s))$ }
    \STATE $s \gets (1-\alpha) s + \alpha \nabla_{w}^2 l$;\quad $\mu \gets \mu - \alpha s^{-1} \nabla_{w} l $ 
\ENDFOR
\STATE return $(\mu, s)$
\end{algorithmic}
\end{algorithm}

\paragraph{ROBUST HMC}
A similar modification needs to be made in the case of Hamiltonian Monte Carlo inference.
When computing the potential energy function the same procedure outlined in lines 6-9 is employed. 
Namely, we have $U(w) = - log(p(w)) -  log(\lambda {Y_{\text{clean}}} + (1-\lambda){Y_{\text{worst}}})$, where the vectors ${Y_{\text{clean}}}$ and ${Y_{\text{worst}}}$ are defined as in Algorithm~\ref{alg:RobustVI}. 
Observe that the kinetic function given in Section \ref{sec:background} remains unaltered, as it is not related to the weight distribution.

\section{EXPERIMENTS}\label{sec:experimens}

We conduct an empirical evaluation of our framework considering various approximate inference methods for BNNs, including SWAG \citep{maddox2019simple}, NoisyAdam (NA) \citep{zhang2018noisy}, Variational Online Gauss Newton (VOGN) \citep{khan2018fast}, Bayes by Backprop (BBB) \citep{blundell2015weight}, and Hamiltonian Monte Carlo (HMC) \citep{neal2011mcmc}. 

We first evaluate the robust accuracy of networks trained with our robust likelihood compared to standard training on the MNIST benchmark \citep{mnist}. We then perform analogous analyses on the CIFAR-10 \citep{cifar} dataset, and, finally, we empirically study the effect of our robust likelihood on the predictive uncertainty. Further, training parameters, including prior distribution and architecture for each of the BNNs,  can be found in the Supplementary Material.

For HMC, we set the initial weight to be a sample from the prior when performing standard training and set the initial weight to a pre-trained SGD iterate when performing inference with robust likelihood; this is to enforce that the starting point of the algorithm is closer to the target distribution.


\paragraph{EVALUATION ON MNIST}

We use $p_\epsilon$ introduced in Eqn \eqref{Eqn:discreteDistr} with $\eta = 0.1$ and $\lambda = 0.25$ (an empirical evaluation of the effect of changing these parameters, as well as using different forms for $p_\epsilon$, is reported in the Supplementary Material). We train a single hidden layer BNN with 512 neurons on the full MNIST dataset (further training parameters are given in the Supplementary Material).

\begin{figure}[h]
\centering
\includegraphics[width=0.425\textwidth]{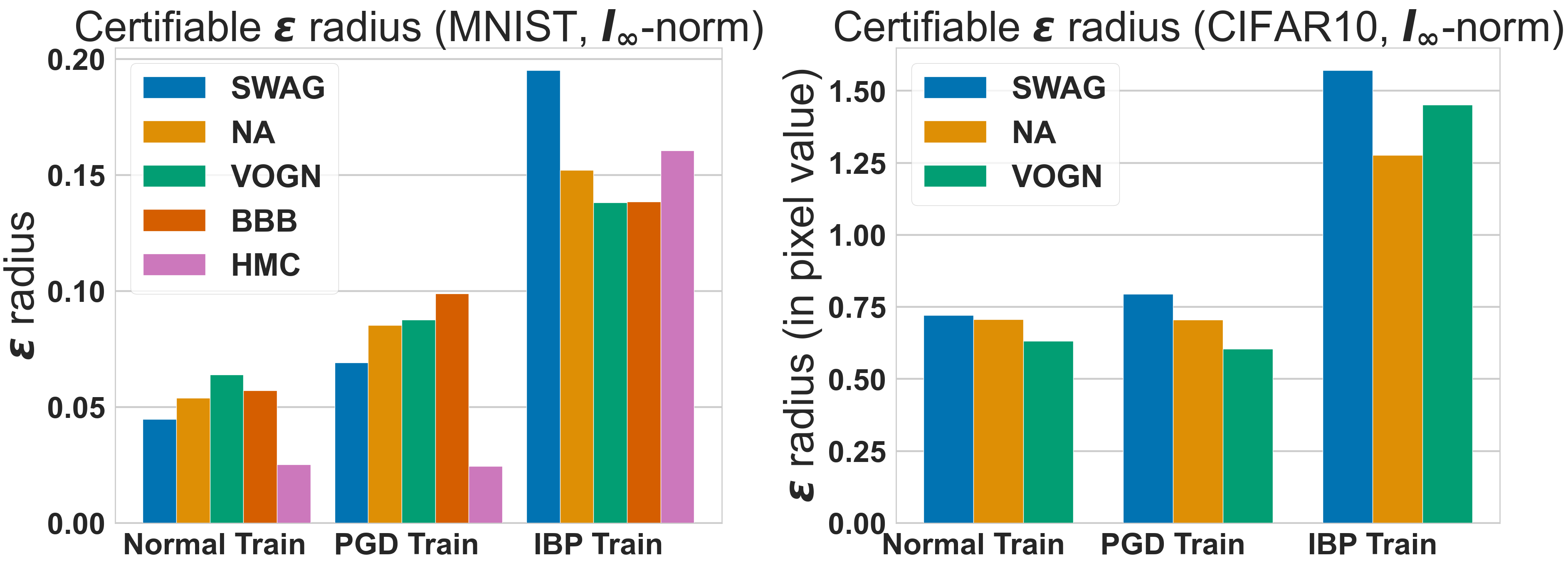}
\caption{We plot the average certified radius for images from MNIST (right), and CIFAR-10 (left) using the methods of \cite{boopathy2019cnn}. We observe that robust training with IBP roughly doubles the maximum verifiable radius 
compared with standard training and that obtained by training on PGD adversarial examples.}\label{fig:CNN-Cert}
\end{figure}

In Figure~\ref{fig:MNIST} (top row) we analyze how different training methods affect the accuracy, robustness to PGD attacks ($\mathcal{R}_{0.1}^{PGD}$), as well as the certified lower bounds using IBP ($\mathcal{R}_{0.1}^{IBP}$) and LBP ($\mathcal{R}_{0.1}^{LBP}$). We use both IBP as well as the more computationally expensive but tighter LBP in order to study the effect of training with our robust likelihood without the bias of training and evaluating with the same certification method. Moreover, in Figure~\ref{fig:CNN-Cert} we analyze the maximum certifiable radius (via the methodology in \citep{boopathy2019cnn}) in order to further compare the effects of each likelihood.


In Figure~\ref{fig:MNIST}, we find that,  while all BNNs trained with the standard likelihood (left plot) perform comparably well in terms of accuracy, there is a marked difference in their robustness against PGD. This is in line with what was observed by \cite{carbone2020robustness}, where the more fidelity an inference method has to the true Bayesian posterior, the greater is its robustness to gradient-based attacks. However, the certified robust accuracy obtained using standard likelihood is identically zero, that is, we obtain no certification in these settings. This implies, for example, that although HMC is resistant to PGD attacks, we cannot guarantee that a different, successful attack method does not exist.
Similarly, for training with the PGD approximation of the robust likelihood (central plot), we obtain that, while the robustness against PGD of each model is now around $80\%$, the certified robustness is still at $0\%$. Similar behaviour has also been observed for adversarial training with gradient-based attacks for deterministic neural networks \citep{gowal2018effectiveness}. 
By using IBP during training to lower bound the robust likelihood 
(right plot in the figure), we find that, not only do we obtain similar levels of accuracy and PGD robustness as before, but we are also able to provide non-trivial certification on the robust accuracy, $\mathcal{R}_\epsilon$, of the networks, that is, against \textit{any} possible adversarial perturbation of magnitude up to $\epsilon = 0.1$.
For example, using SWAG we obtain $\mathcal{R}_{0.1} \approx 75\%$, that is, the BNN trained with SWAG with our robust likelihood is provably  adversarially robust on $75\%$ of the points included in the MNIST dataset.

\paragraph{Evaluation on FashionMNIST}

In the center row of Figure~\ref{fig:MNIST}, we use the same networks, $p_{\epsilon}$ distribution (with $\eta = 0.1$ and $\lambda = 0.25$) and evaluation methods stated for MNIST, but applied to the FashionMNIST dataset \citep{xiao2017fashion}. Despite being a harder dataset than MNIST, as evidenced by the reduced accuracy of the approximate posteriors, we find the robustness trends to be qualitatively similar to those on MNIST and CIFAR10. We do note, however, that PGD training was much more effective at increasing the certified bound when the bound is computed with LBP. Our hypothesis about this difference is that because the networks are less accurate they are also potentially less robust to gradient-based attacks (following the logic of \citep{carbone2020robustness}) and thus, PGD is able to find strong adversarial examples which can increase the robustness of the posterior. 


\begin{figure*}[ht]
\centering
\includegraphics[width=0.90\textwidth]{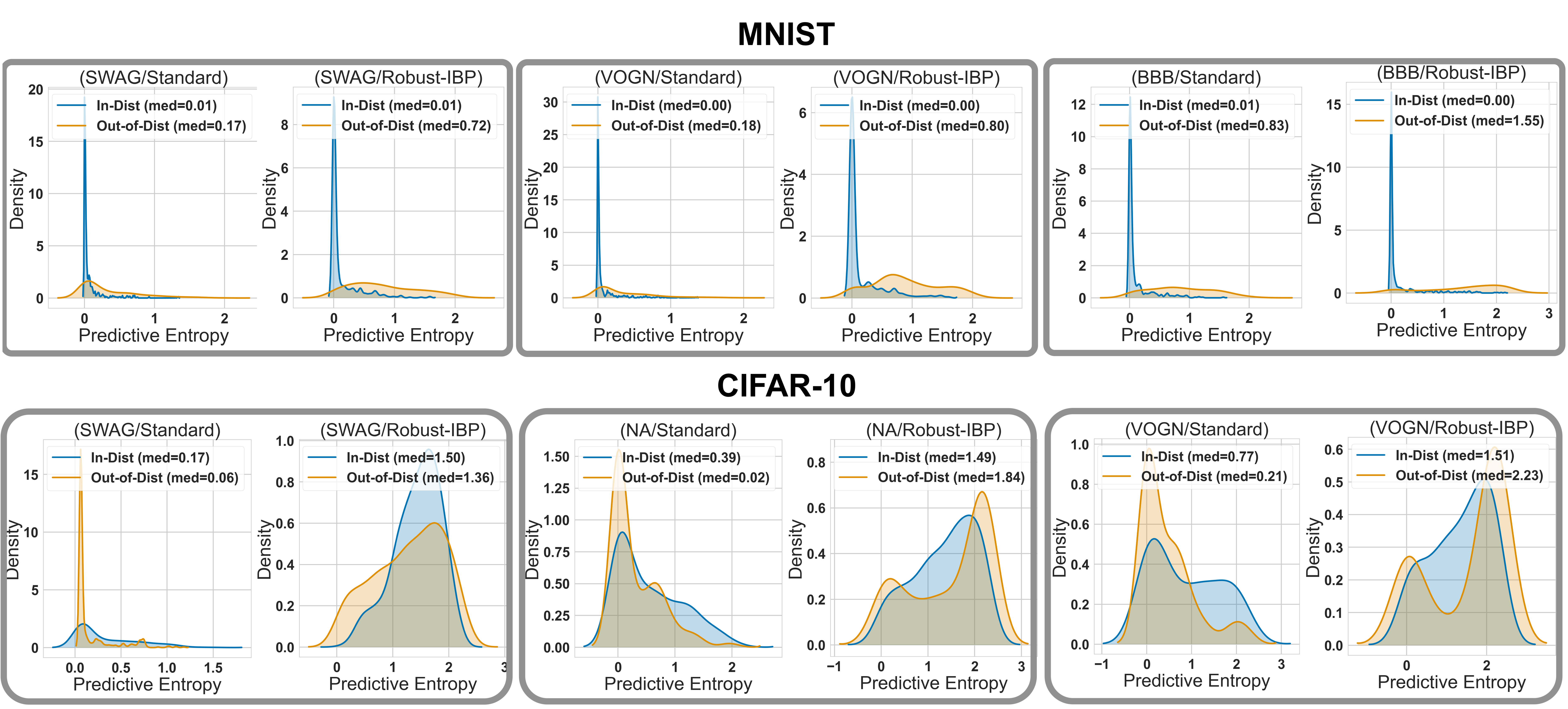}
\caption{We plot the in-distribution (blue) and out-of-distribution (orange) predictive uncertainty. Each pair of plots corresponds to an inference method where the left plot represents the entropy distribution for standard training and the right plot represents robust IBP training. For both MNIST (top row) and CIFAR-10 (bottom row) we find that robust training improves the uncertainty calibration of the network w.r.t. out-of-distribution samples.}\label{fig:entropy}
\end{figure*}

\paragraph{EVALUATION ON CIFAR-10}
We now evaluate the effect of the robust likelihood on BNNs trained on the CIFAR-10 dataset.
The 
CIFAR-10 is more challenging compared to MNIST, and hence not all the training methods considered for MNIST can be used to train reasonably accurate BNNs on this dataset \citep{blier2018description}. 
Consequently, for CIFAR-10 we provide results only for SWAG, NA and VOGN.
In particular, we train a Bayesian convolutional neural network (CNN) with 2 convolutional layers (each with 32, 4 by 4 filters) followed by a max pooling layer and  2 fully connected layers (one with 512 hidden neurons and the other with 10).
For the robust likelihood, we consider Eqn~\eqref{Eqn:discreteDistr} with  $\eta = 1/255$ and $\lambda = 0.25$. Finally, we introduce the standard exponential decay on the learning rate to ensure stable convergence (results for additional parameter values are given in the Supplementary Material).

We perform a similar evaluation to that discussed for MNIST, the results of which are plotted in the bottom row of Figure \ref{fig:MNIST}.
Consistently with what we observed for MNIST, we obtain that BNNs trained by using the standard likelihood (left plot) and PGD attacks (central plot) do not allow for the computation of certified guarantees (the lower bound of the bars is close to zero for all the inference methods). 
In contrast, for the BNN trained with our robust likelihood and IBP we find that, even for CIFAR-10, we are able to compute non-trivial lower bounds on $\mathcal{R}_{\epsilon}$.
For instance, on SWAG we obtain  $\mathcal{R}_{\epsilon}^{IBP}\approx 50\%,$ which is comparable to state-of-the-art results with adversarial training of deterministic NNs on CIFAR-10 \citep{boopathy2019cnn}. 
We analyse the effect that robust training at a specific $\epsilon$ has on the robustness of the network at other values of $\epsilon$.
In order to do so, we employ the method developed by \cite{boopathy2019cnn}, which, for each test image, computes the maximal adversarial perturbation, $\epsilon$, such that the image is provably safe in the related $\epsilon-$ball. 
The results of this analysis are given in Figure~\ref{fig:CNN-Cert}, where we report the average maximum adversarially safe radius over 100 test CIFAR-10 images.
%
We find that, while PGD training does not increase the robustness of the model compared to standard training, our training procedure is able to roughly double the robustness for the three training methods explored here (further results on MNIST are in the Supplementary Material).

\paragraph{EVALUATION ON OUT-OF-DISTRIBUTION UNCERTAINTY}

One of the main features of BNNs, which distinguishes them from DNNs, is their ability to model their prediction's uncertainty \citep{neal2012bayesian}. 
We investigate how training with our robust likelihood influences uncertainty calibration when testing on data that are outside of the training data distribution. 
As is common in the literature \citep{maddox2019simple, lakshminarayanan2017simple}, we evaluate uncertainty on in-distribution and out-of-distribution images using the entropy of the posterior predictive distribution. 
For $x\in \mathbb{R}^n$ the latter is defined as  $\mathbb{E}_{w\sim p(w|\mathcal{D})}[-\sum_{c =1}^C \sigma_c(f^{w}(x)) \log (\sigma_c(f^{w}(x)))]$. 
In particular, we expect that the maximal class probabilities on out-of-distribution images will have high entropy, reflecting the model’s uncertainty in its predictions, and considerably lower entropy on images that are similar
to those on which the network was trained (in-distribution). 
In the top row of Figure~\ref{fig:entropy}, we plot the entropy of the posterior predictive distributions on in-distribution test samples (in blue) and on FashionMNIST samples (in orange), which are out-of-distribution data for networks trained on the MNIST dataset. 
For CIFAR-10 (bottom row in the figure) we use the SVHN \citep{goodfellow2013multi}  dataset as the source of meaningful, but out-of-distribution, data. We hypothesize that if the robust likelihood introduces meaningful information about invariances in the data, then the uncertainty of the resulting posterior would be improved. 
On MNIST, we find that, in each case, robust training with IBP significantly improves uncertainty by making more uncertain predictions on out-of-distributions data, while still being confident on in-distribution data. 
For CIFAR-10, surprisingly, we find that BNNs trained with the standard likelihood  are more confident on the out-of-distribution data than on in-distributions data, a behaviour that, with the exception of SWAG, is always reversed by performing robust training with IBP. 
In all the cases, training with IBP on CIFAR-10 increases the entropy of the predictions, which may be indicative of better calibrated uncertainty compared to the normally inferred models (note that on CIFAR-10 accuracy is of the order of $60\%$, and hence we do not expect the model to be very confident even on in-distribution images).
In the Supplementary Materials, to further confirm our empirical results, we evaluate the entropy on adversarial examples and also analyze the likelihood ratio between in-distribution and out-distribution data. 

\section{CONCLUSION}
We presented a framework for robust training of certifiably robust deep neural networks based on a principled Bayesian foundation. We developed an algorithmic implementation of our framework that employs constraint relaxation and can be integrated with existing approximate inference methods for BNNs. 
On the MNIST, FashionMNIST and CIFAR-10 datasets we showed that, not only does our framework allow us to train BNNs that are guaranteed to be robust to adversarial examples, it can also have a positive effect on uncertainty calibration.

\paragraph{Acknowledgements}

This project was funded by the ERC under the European Union’s Horizon 2020 research and innovation programme (FUN2MODEL, grant agreement No.~834115). ZC and ZZ are supported by DOE grant No. DE-SC0021323. 

\bibliography{paper}

\onecolumn
\aistatstitle{Supplementary Materials: Bayesian Inference with Certifiable Adversarial Robustness}

In these supplementary materials we provide the details to aid in the reproducibility of our results and report on further experiments to deepen our understanding of the presented method. For the code to reproduce both the experiments found in the main text and in these extended materials see the github code repository at: \textit{https://github.com/matthewwicker/CertifiableBayesianInference}, and if anything proves to be unclear or broken please email Matthew Wicker at: \textit{matthew.wicker@cs.ox.ac.uk}.

\section{APPROXIMATE INFERENCE PARAMETERS}

In this section of the Supplementary Material, we list the training parameters that we used for the training of each of the networks discussed in the main text.


\subsection{MNIST and FashionMNIST Parameters}
\begin{table}[ht]
\centering
\begin{tabular}{@{}|l|l|l|l|l|l|@{}}
\toprule
               & SWAG   & NoisyAdam & VOGN     & BBB      & HMC      \\ \midrule
Learning Rate  & 0.1    & 0.001     & 0.35     & 0.45     & 0.075    \\
Prior Scaling  & N/A    & 10         & 10        & 20       & 500      \\
Batch Size     & 128    & 128       & 128      & 128      & 60k      \\
Epochs/Samples & 20/250 & 20/(N/A)  & 20/(N/A) & 20/(N/A) & (N/A)/25 \\
PGD Iterations & 10     & 10        & 10       & 10       & 10       \\ \bottomrule
\end{tabular}
\end{table}

Each network trained on MNIST is a single hidden layer fully-connected architecture with 512 neurons in the hidden layer.
The parameters used for the 5 training methods are listed in the table above.
Prior scaling refers to a multiplicative constant w.r.t.\ the initialisation parameters described in \cite{sutskever2013importance}.  
In fact, we often find the initial variance described in the later to be  too small for retrieving good uncertainty estimates, and, thus, we further multiply it by the values reported in the table.
Further parameters that are specific to HMC, and not included in the table, are: 3 iterations of burn-in, with 20 steps of the leapfrog numerical integrator followed by the reported 25 samples from the posterior each which explore the chain for 25 steps with the leapfrog integrator. 
We again note that when we perform approximate inference with HMC and the robust likelihood that we choose the initial network parameters to be the result of 10 epochs of stochastic gradient descent rather than the full-data gradient descent used during normal burn-in. 
Finally, we note that we follow the empirically optimal procedure stated by \cite{gowal2018effectiveness}. 
In particular, we train with an $\eta$ linearly increasing to its target value at every epoch. 
Again as in \cite{gowal2018effectiveness}, we set the target $\eta$ value 10\% larger than the `desired' robustness value.

\subsection{CIFAR10 Parameters}
\begin{table}[ht]
\centering
\begin{tabular}{@{}|l|l|l|l|@{}}
\toprule
               & SWAG   & NoisyAdam & VOGN     \\ \midrule
Learning Rate  & 0.015  & 0.00025   & 0.25     \\
LR Decay       & 0.0    & 0.025     & 0.025    \\
Prior Scaling  & N/A    &   5       &   5      \\
Batch Size     & 128    & 128       & 128      \\
Epochs/Samples & 45/500 & 45/(N/A)  & 45/(N/A) \\
PGD Iterations & 10     & 10        & 10       \\ \bottomrule
\end{tabular}
\end{table}

For CIFAR10, prior to inference we perform data augmentation which involves horizontal flipping as well as random translations by up to 4 pixels. We randomly select an image from the train set with uniform probability and then select a transformation (translation or horizontal flipping) until we have augmented the data size from 60k to 100k images 
Finally, the network architecture is made of two convolutional layers, respectively with 16 and 32 four by four filters, followed by a 2 by 2 max pooling layer, and  a fully connected layer with 100 hidden neurons.

\section{CERTIFIED ROBUST RADIUS RESULTS}

\begin{figure*}[ht]
\vspace{-1em}
\centering
\includegraphics[width=1.0\textwidth]{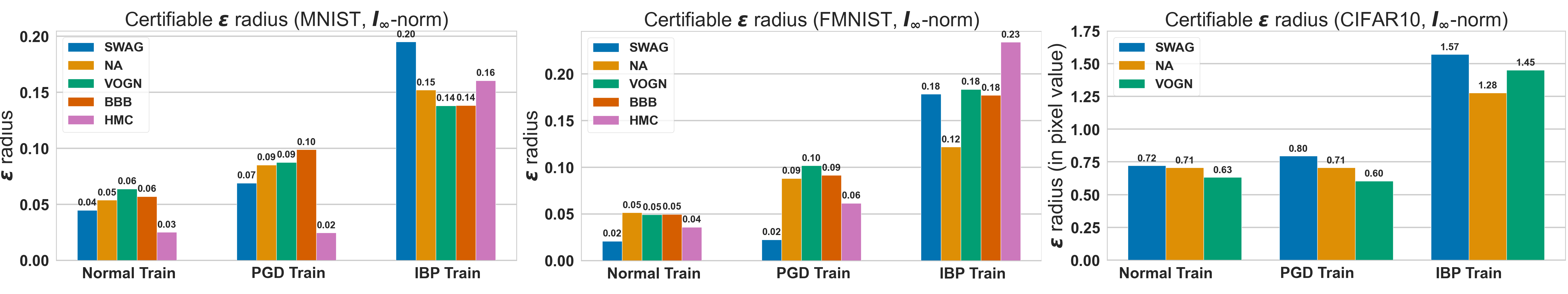}
\caption{We plot the average certified radius for images from MNIST (left), FashionMNIST (center), and CIFAR10 (right) using the methods of \cite{boopathy2019cnn}. We re-report the MNIST and CIFAR10 results here for ease of comparison. We observe that robust training with IBP roughly doubles the maximum verifiable radius of compared with standard training and that obtained by training on PGD adversarial examples. }\label{fig:saferad}
\end{figure*}

Consistent with the analysis in the main text, we consider analyzing the robustness of the trained posteriors at varying values of $\epsilon$ (reported in Figure~\ref{fig:saferad}). In particular, we estimate the maximal $\epsilon$ radius for which each image is robust. To estimate this value, we follow the methodology of \cite{boopathy2019cnn}: a binary search over the values of $\epsilon$. We stress that during this procedure, we use linear propagation methods: CROWN for MNIST and FMNIST networks and CNN-Cert for CIFAR10 Networks. This is to reduce the bias of the evaluation of IBP trained networks. That is, IBP trained networks intuitively should evaluate well against IBP but it is important to see if tighter methods still show large improvements. As reported in the paper, we find that training with PGD does not tend to increase the certifiable radius in a significant way, while training with IBP allows one to double the certifiable radius.


\section{ADVERSARIAL TRAINING PARAMETER STUDY}

\begin{figure*}[ht]
\centering
\includegraphics[width=\textwidth]{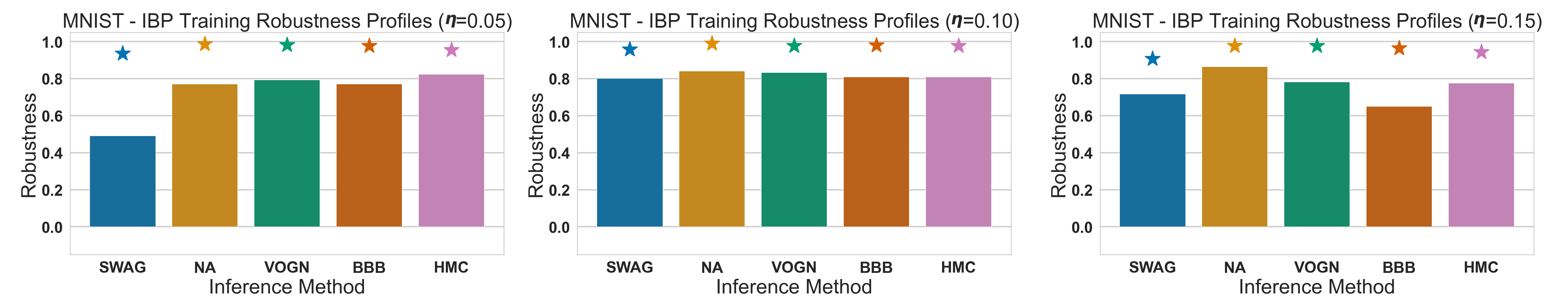}
\includegraphics[width=\textwidth]{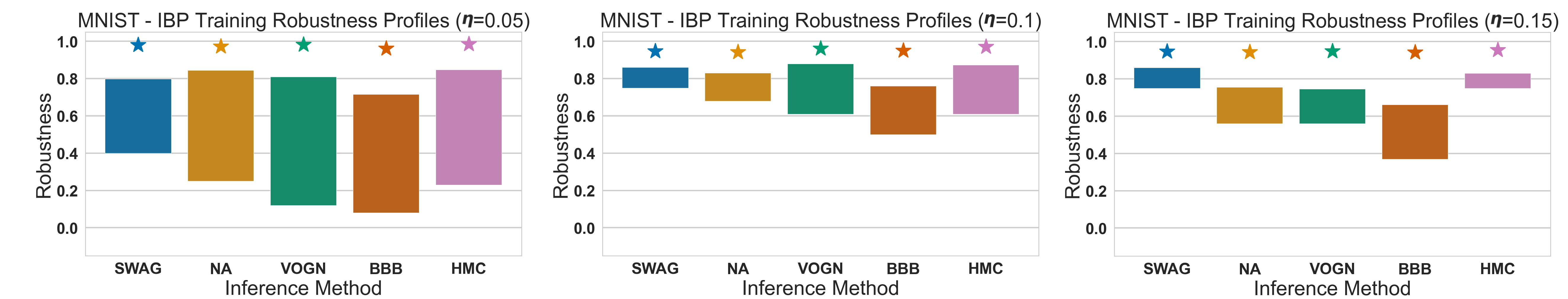}
\caption{\textbf{Left to Right:} Effect of varying (increasing) values of $\eta$ on the robustness profile of resulting approximate posteriors. \textbf{Top Row:} Robustness profiles of networks using the robust likelihood with PGD as an approximate worst-case adversary. \textbf{Bottom Row:} Robustness profiles of networks using the robust likelihood with IBP as an approximate worst-case adversary. Accuracy (plotted as star points), an empirical estimation of $\mathcal{R}_{\epsilon}$ obtained using PGD (upper bound of each bar), and $\mathcal{R}_{\epsilon}^{IBP}$ (lower bound of each bar), obtained for $\epsilon = 0.1$ on the MNIST dataset.}\label{fig:eta}
\end{figure*}

\begin{figure*}[ht]
\centering
\includegraphics[width=\textwidth]{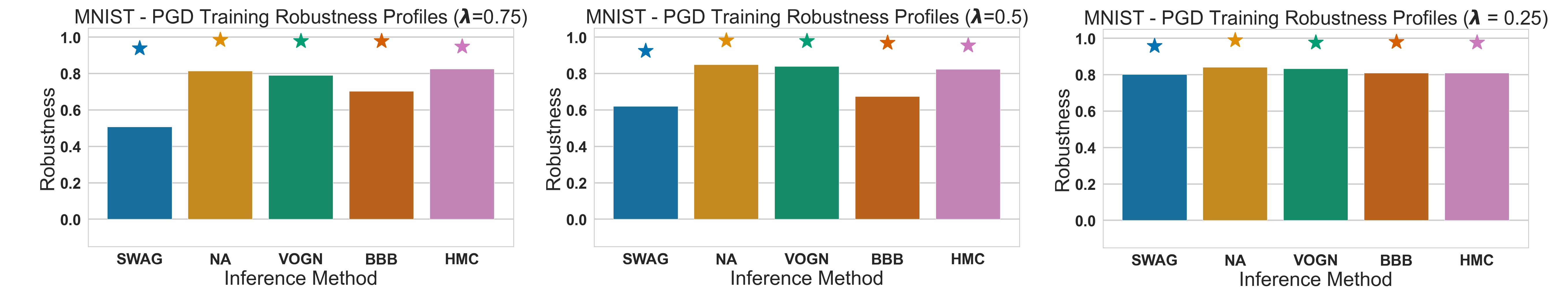}
\includegraphics[width=\textwidth]{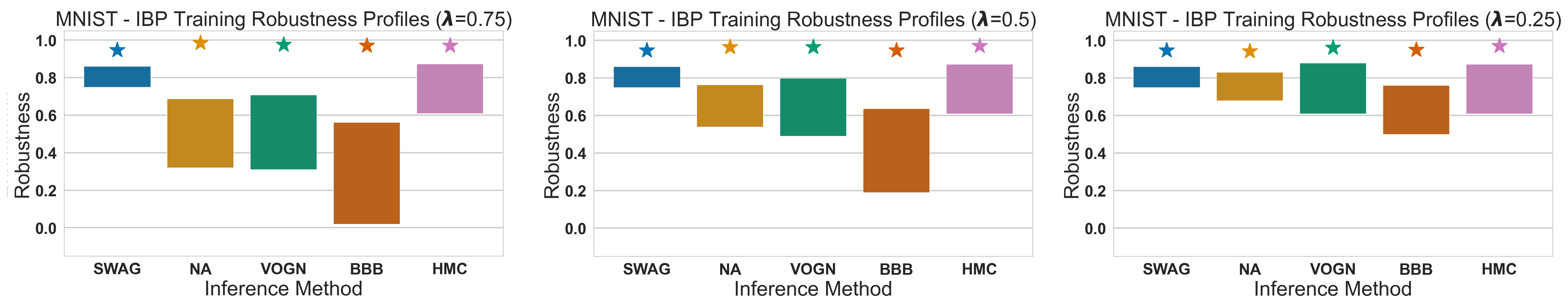}
\caption{ \textbf{Left to Right:} Effect of varying (decreasing) values of $\lambda$ on the robustness profile of resulting approximate posteriors. \textbf{Top Row:} Robustness profiles of networks using the robust likelihood with PGD as an approximate worst-case adversary. \textbf{Bottom Row:} Robustness profiles of networks using the robust likelihood with IBP as an approximate worst-case adversary. Accuracy (plotted as star points), an empirical estimation of $\mathcal{R}_{\epsilon}$ obtained using PGD (upper bound of each bar), and $\mathcal{R}_{\epsilon}^{IBP}$ (lower bound of each bar), obtained for $\epsilon = 0.1$ on the MNIST dataset.}\label{fig:lambda}
\end{figure*}

In this section we analyse the choice of $p_\epsilon$, that is, the distribution that controls the adversarial perturbation strength at training time.
Recall that the distribution used in the main text follows related work on training of deterministic neural networks:
\begin{align}
    p_{\epsilon}(\epsilon)=\begin{cases}
    \lambda      & \quad \text{if } \epsilon=0\\
    1-\lambda      & \quad \text{if } \epsilon=\eta
  \end{cases}.
  \label{Eqn:discreteDistr}
\end{align}
In particular, we first study the affect of changing the $\lambda$ parameter in Eqn~\eqref{Eqn:discreteDistr} which parameterizes the relative penalty between accuracy and robustness during inference. Next, we study the effect of changing $\eta$ in Eqn~\eqref{Eqn:discreteDistr} which sets a the maximum allowable manipulation magnitude during inference.  Finally, we study the effect of changing the form of the $\epsilon$ probability density function to two different continuous, non-negative distribution. 

In each figure, we maintain the plotting conventions used in Figure 1 of the main text.
For each posterior: accuracy is plotted as a star point, an empirical estimation of $\mathcal{R}_{\epsilon}$ obtained using PGD (upper bound of each bar), and $\mathcal{R}_{\epsilon}^{IBP}$ (lower bound of each bar), obtained for $\epsilon = 0.1$ on the MNIST dataset.  For the following analysis we only report the lower-bound based on IBP. 

\subsection{The Effect of Adversarial Magnitude During Inference}

When approximating the robust likelihood with PGD during inference, we find that the shift in magnitude of $\eta$ on the resulting robustness estimates is largely dependant on the method of approximate inference. Interestingly, we find that for SWAG and BBB, that training with $\eta = 0.15$ becomes problematic as it seems that with the current training parameters (reported in the previous section), the 1 layer, 512 neuron network may not have had enough capacity to accurately capture good adversarial robustness. The connection between robustness of (gradient-based) adversarial trained deterministic networks and capacity is discussed at length in \cite{madry2017pgd}. We find that NA and HMC are relatively unaffected by small changes to the $\eta$ magnitude and enjoy similar heightened robustness for each observed value. 

The effect of $\eta$ is much more pronounced when we perform inference with the IBP robust likelihood. We see that having an $\eta$ smaller than $\epsilon$ (in $\mathcal{R}_{\epsilon}$)  results in worse lower-bound potentially indicating a less robust posterior. For parameter and natural gradient VI, we also find that having an $\eta$ that is much larger than $\epsilon$ can be detrimental as too strong of an adversary can be problematic for learning.

\subsection{The Effect of Trading Accuracy and Robustness}

In Eqn~\eqref{Eqn:discreteDistr} the parameter $\lambda$ effectively controls the relative weighting of accuracy-error and robust-error during the inference procedure. Specifically, we note the cases $\lambda = 1.0$ which results in the standard likelihood (a.k.a.\ the categorical cross-entropy in the case of classification), and $\lambda = 0.0$ results in a framework in which give importance solely to robustness. 
In Figure~\ref{fig:lambda} we report the change in robustness profiles for $\lambda \in \{0.75, 0.5, 0.25\}$ for training with the worst-case approximated by PGD (top row) and IBP (bottom row). 

When approximating the robust likelihood with PGD, we find that HMC and natural gradient methods (VOGN, NA) are not strongly affected by the choice of $\lambda$, whereas we see the most pronounced difference with SWAG which is greatly affected by choice of $\lambda$. In particular we highlight roughly a $20\%$ raw increase in the robustness to gradient based attacks for each 0.25 decrease in $\lambda$. 
On the other hand, when training with IBP there is large shift in the resulting robustness profiles for parameter and natural gradient VI methods (BBB, VOGN, NA). Notably, we see a large (50\% raw) increase in the lower-bound for BBB as the value for $\lambda$ varies between 0.75 and 0.25.

\subsection{On the Choice of Density Function for Adversarial Magnitude}

\begin{figure*}[ht]
\centering
\includegraphics[width=\textwidth]{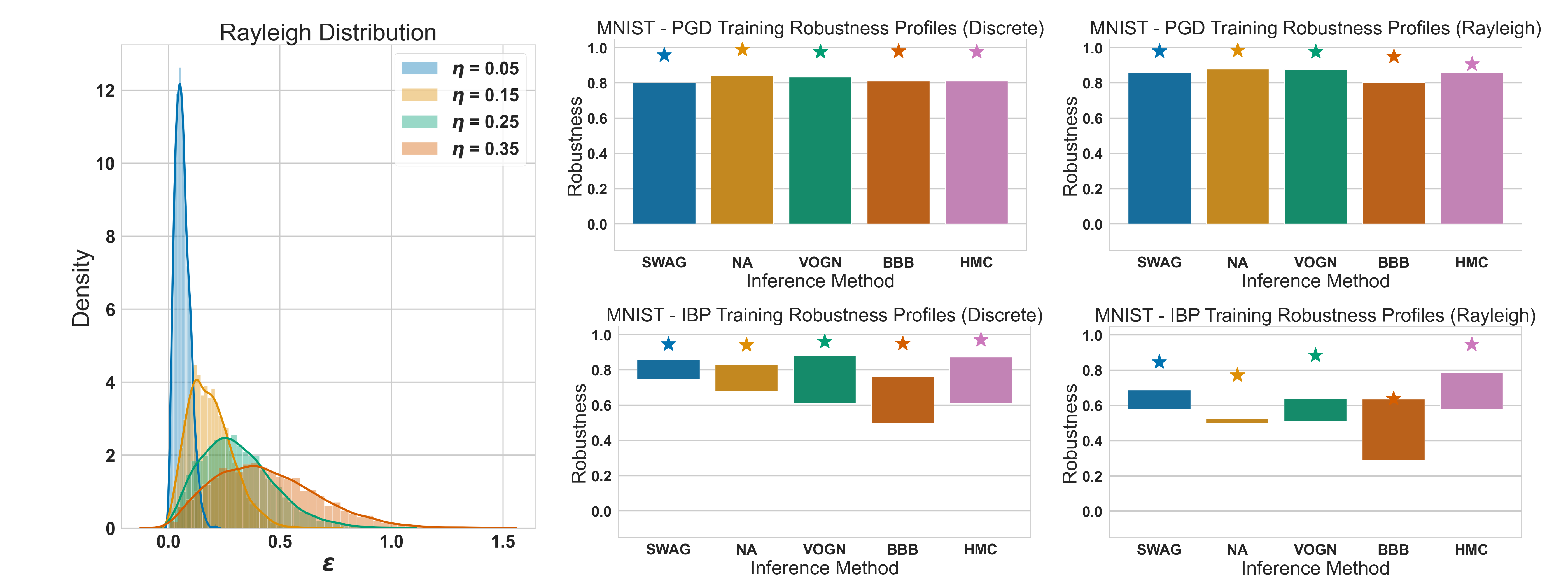}
\caption{ \textbf{Left:} Effect of varying the scale $\eta$ of the Rayleigh distribution on the density $p_\epsilon$ when training we use $\eta = 0.1$. \textbf{Right, Top Row:} Robustness profiles of networks using the robust likelihood with PGD as an approximate worst-case adversary. \textbf{Right, Bottom Row:} Robustness profiles of networks using the robust likelihood with IBP as an approximate worst-case adversary. Accuracy (plotted as star points), an empirical estimation of $\mathcal{R}_{\epsilon}$ obtained using PGD (upper bound of each bar), and $\mathcal{R}_{\epsilon}^{IBP}$ (lower bound of each bar), obtained for $\epsilon = 0.1$ on the MNIST dataset.}\label{fig:pepsdist-ray}
\end{figure*}

\begin{figure*}[ht]
\centering
\includegraphics[width=\textwidth]{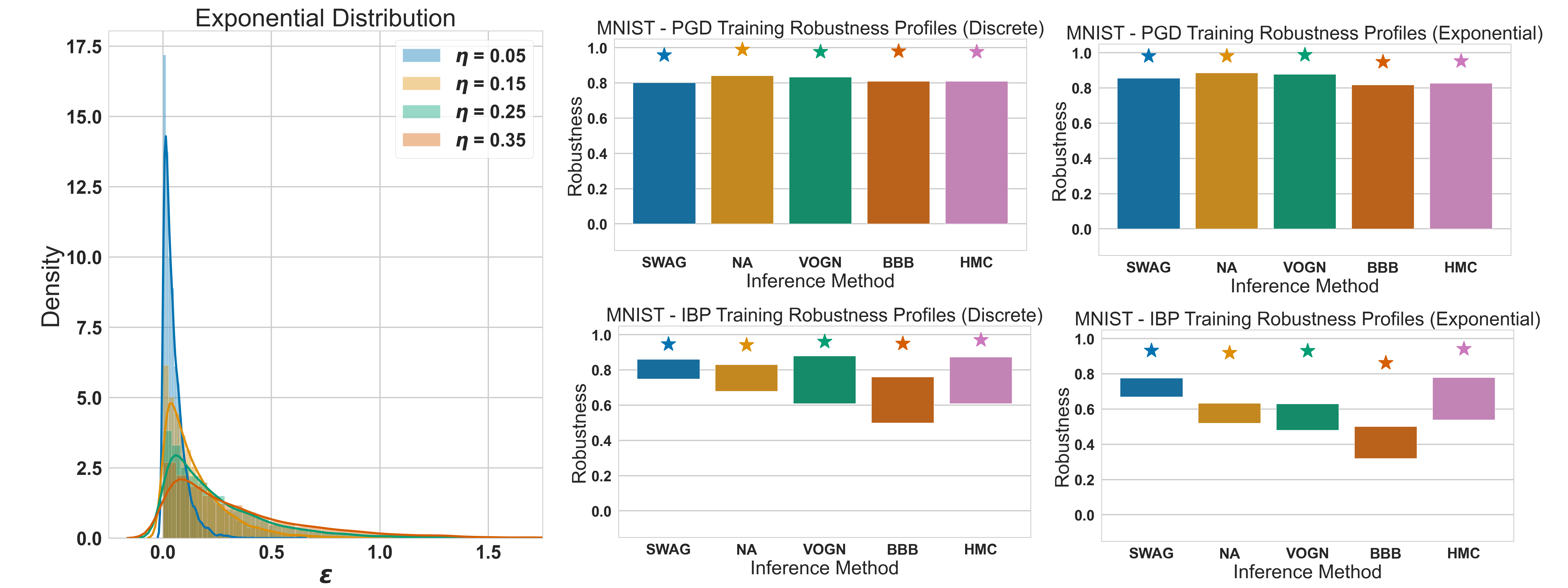}
\caption{\textbf{Left:} Effect of varying the scale $\eta$ of the exponential distribution on the density $p_\epsilon$ when training we use $\eta = 0.1$. \textbf{Right, Top Row:} Robustness profiles of networks using the robust likelihood with PGD as an approximate worst-case adversary. \textbf{Right, Bottom Row:} Robustness profiles of networks using the robust likelihood with IBP as an approximate worst-case adversary. Accuracy (plotted as star points), an empirical estimation of $\mathcal{R}_{\epsilon}$ obtained using PGD (upper bound of each bar), and $\mathcal{R}_{\epsilon}^{IBP}$ (lower bound of each bar), obtained for $\epsilon = 0.1$ on the MNIST dataset.}\label{fig:pepsdist-exp}
\end{figure*}


In Figure~\ref{fig:pepsdist-ray} and Figure~\ref{fig:pepsdist-exp}, we study changing the form of $p_\epsilon$ from the density given in Eqn~\eqref{Eqn:discreteDistr} to a Rayleigh distribution and an Exponential distribution, respectively. We have chosen these distributions in particular because they have non-negative support and a single controlling variable. In principle, however, any distribution (with a positive support) can be chosen for the form of $p_\epsilon$. As noted in the main text, during the computation of the loss function, one must marginalize over the selected $p_\epsilon$ distribution, which in this case is done via Monte Carlo with only 10 samples from $p_\epsilon$ per batch. Consistent with the study presented in the main text, we evaluate robustness profiles with $\epsilon$ set to 0.1.

\subsubsection{Using a Rayleigh Distribution}

In Figure~\ref{fig:pepsdist-ray}, we plot the case in which training is done by using an Rayleigh distribution with the scale set to $\eta$ for $p_\epsilon$ as follows:
\begin{align}
     p_{\epsilon}(\epsilon) = \dfrac{\epsilon}{\eta^2} \text{exp}\bigg(\dfrac{-\epsilon^2}{2\eta^2}\bigg)
\end{align}
In our experiments, we find that using a Rayleigh distribution for $p_\epsilon$ does marginally improve the robustness ($\mathcal{R}_{\epsilon}$) when training against a PGD adversary ($\approx 4\%$ on average). We also find that when using the altered pdf, the main result stated in the paper, that training with robust likelihood is the only method that gives non-trivial lower bounds on robustness, still holds. However, we find that the use of the Rayleigh distribution has an adverse affect on the overall robustness profile compared to training with Eqn~\eqref{Eqn:discreteDistr}. 

\subsubsection{Using an Exponential Distribution}

In Figure~\ref{fig:pepsdist-exp}, we give the results when $p_\epsilon$ is selected as an exponential distribution with the rate set to $\eta^{-1}$:
\begin{align}
     p_{\epsilon}(\epsilon) = \dfrac{1}{\eta}\text{exp}\left(\dfrac{-\epsilon}{\eta}\right)
\end{align}
When training against a PGD adversary, we found that an using an exponential distribution for $p_\epsilon$ also leads to small increases in robustness against adversarial attacks, with an average increase of $\approx 5\%$. Consistent with the results for the Rayleigh distribution, the main result stated in the paper, that training with the robust likelihood  is the only method that gives non-trivial lower bounds on robustness. However, we continue to find that the use of the exponential distribution when training with IBP, consistent with the Rayleigh distribution, has an adverse affect on the overall robustness profile compared to training with Eqn~\eqref{Eqn:discreteDistr}.

\begin{figure*}[ht]
\centering
\includegraphics[width=0.55\textwidth]{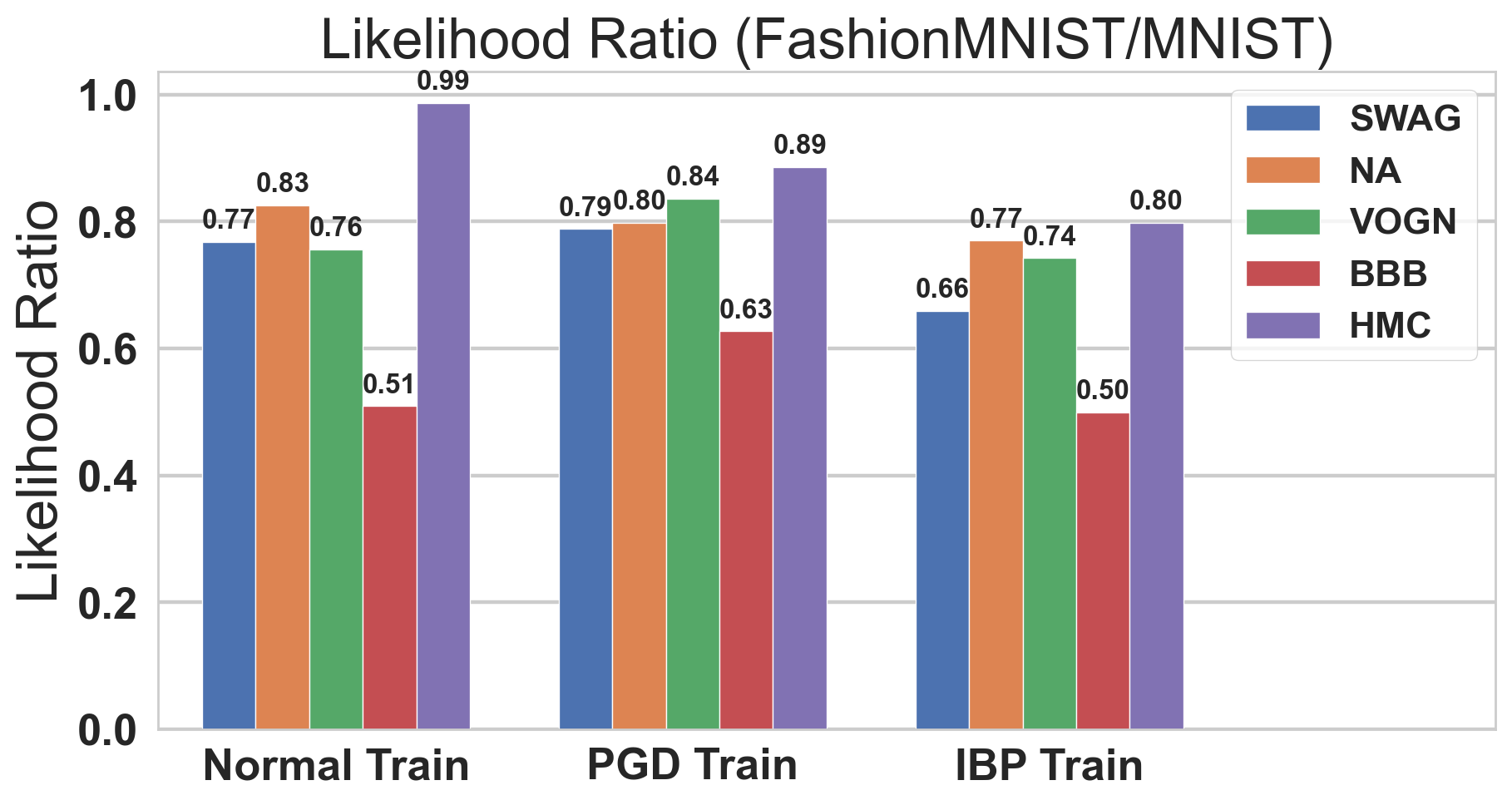}
\caption{Likelihood Ratios using FashionMNIST as out-of-distribution samples for posteriors inferred on the MNIST datset. A likelihood ratio of 1.0 represents predictions which are equally confident in and out of distribution. Conversely, a low likelihood ratio represents less certain predictions on out-of-distribution points. }\label{fig:llr}
\end{figure*}

\begin{figure*}[ht]
\centering
\includegraphics[width=\textwidth]{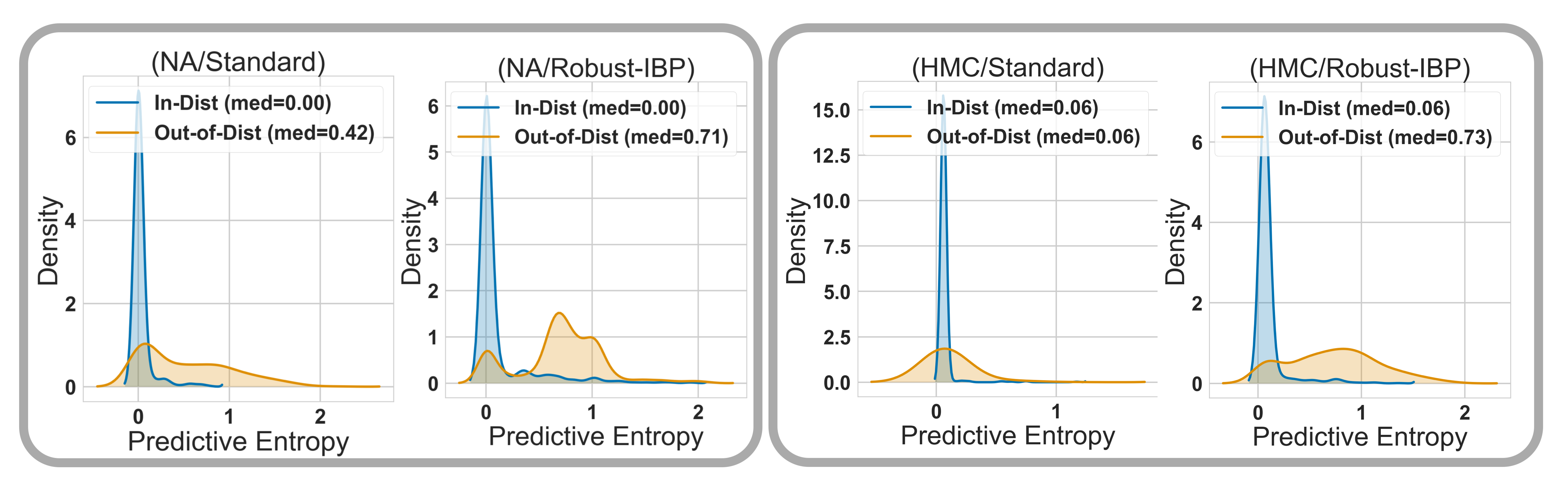}
\caption{We plot the in-distribution (blue) and out-of-distribution (orange) predictive uncertainty. Each pair of figures corresponds to an inference method where the left figure represents the entropy distributions for standard training and the right figure represents robust IBP training. We find that robust training improves the uncertainty calibration of the network w.r.t. out-of-distribution samples.}\label{fig:ent}
\end{figure*}

\section{Likelihood Ratios}

Expanding briefly on the evaluation of uncertainty on out-of-distribution points, we also observe the affect of training with robust likelihood on the `likelihood ratio' of in and out-of-distribution (OOD) points. Similarly to how we evaluate OOD points in the main text, we use FashionMNIST dataset as out-of-distribution points for networks trained on MNIST. The likelihood ratio is calculated as the average softmax probability coming from out-of-distribution points divided by the average predictive probability of in-distribution points. Thus, a likelihood ratio of 1.0 represents predictions which are equally confident in and out of distribution. Conversely, a low likelihood ratio represents less certain predictions on out-of-distribution points. We show that for our method of training, consistently with the often used entropy measure reported in the main text, IBP training consistently improves the calibration of uncertainty on out-of-distribution points compared with normal training. 

\subsection{Extended Out-of-Distribution Entropy Plots}

In Figure~\ref{fig:ent} we extend the out of distribution MNIST plots given in the main text to the other approximate inference techniques and find that the same result that is discussed in the main text holds for HMC and NA as well. 


\end{document}